# How Ready Are Generative Pre-trained Large Language Models for Explaining Bengali Grammatical Errors?

Subhankar Maity
IIT Kharagpur, India
subhankar.ai@kgpian.
iitkgp.ac.in

Aniket Deroy
IIT Kharagpur, India
roydanik18@kgpian.
iitkgp.ac.in

Sudeshna Sarkar
IIT Kharagpur, India
sudeshna@cse.iitkgp.ac.in

## ABSTRACT
*Grammatical error correction* (GEC) tools, powered by advanced generative artificial intelligence (AI), competently correct linguistic inaccuracies in user input. However, they often fall short in providing essential *natural language explanations*, which are crucial for learning languages and gaining a deeper understanding of the grammatical rules. There is limited exploration of these tools in low-resource languages such as Bengali. In such languages, *grammatical error explanation* (GEE) systems should not only correct sentences but also provide explanations for errors. This comprehensive approach can help language learners in their quest for proficiency. Our work introduces a real-world, multi-domain dataset sourced from Bengali speakers of varying proficiency levels and linguistic complexities. This dataset serves as an evaluation benchmark for GEE systems, allowing them to use context information to generate meaningful explanations and high-quality corrections. Various generative pre-trained large language models (LLMs), including GPT-4 Turbo, GPT-3.5 Turbo, Text-davinci-003, Text-babbage-001, Text-curie-001, Text-ada-001, Llama-2-7b, Llama-2-13b, and Llama-2-70b, are assessed against human experts for performance comparison. Our research underscores the limitations in the automatic deployment of current state-of-the-art generative pre-trained LLMs for Bengali GEE. Advocating for human intervention, our findings propose incorporating manual checks to address grammatical errors and improve feedback quality. This approach presents a more suitable strategy to refine the GEC tools in Bengali, emphasizing the educational aspect of language learning.

## 1. INTRODUCTION
Generative artificial intelligence (AI) plays a pivotal role in *grammatical error correction* (GEC), a valuable application of natural language processing [7]. GEC serves practical purposes in text proofreading and supports language learning. Recent strides in generative large language models (LLMs), as highlighted by [31, 7], have notably bolstered the capabilities of GEC systems. However, despite Bengali being the $7^{th}$ most spoken language globally

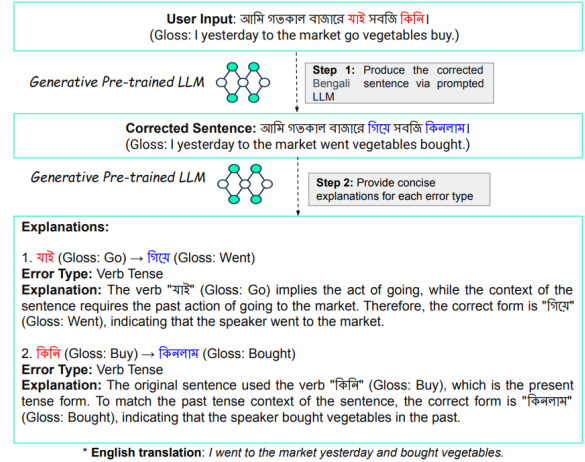

Figure 1: A visual representation of the two-step pipeline of Bengali Grammatical Error Explanation (GEE): Given an erroneous Bengali sentence, the GEE system first produces the corrected sentence. Then, for each corrected error in the sentence, it categorizes the error type and offers a brief explanation. Glosses are provided in round brackets '()'.

[3], current works [23, 27, 19, 18, 1] in Bengali GEC encounter a significant challenge in providing comprehensive explanations for errors in natural language alongside their corrections. The provision of error explanations is pivotal for effective language learning and teaching, as emphasized by [14]. Although corrections deliver implicit feedback, the impact of explicit feedback, which involves identifying errors and offering meta-linguistic insights, such as rules to craft well-structured sentences or phrases, is more substantial, as noted by [13, 15]. In this work, we explore *grammatical error explanation* (GEE) specifically for the Bengali language. This task requires a model to generate natural language error explanations, aiming to assist language learners in acquiring and improving their grammar knowledge. As shown in Figure 1, the GEE system first generates a corrected sentence for a given erroneous Bengali sentence. Subsequently, it categorizes each corrected error in the sentence, providing a brief explanation for each error type. This task aims to address the challenge mentioned above by focusing on the explicit communication of grammatical rules and linguistic insights in the context of error correction. As we have witnessed the versatility of generative LLMs in various tasks [8], in this work, we aim to explore the capabilities of generative pre-trained LLMs, including GPT-4 Turbo [22], GPT-3.5 Turbo, Text-davinci-003, Text-babbage-001 [5], Text-curie-001, Text-ada-001, Llama-2-7b [30], Llama-2-13b and Llama-2-70b, in GEE for

low-resource languages like Bengali. The intricacies of the Bengali language, marked by its *complex morphology*, *diverse verb forms*, and *intricate sentence structures*, present formidable hurdles in Bengali natural language processing [26, 18]. Therefore, the utilization of these LLMs in GEE for Bengali presents a unique set of challenges and opportunities that we seek to investigate further. In the realm of GEE, the exploration of GEC systems in Bengali is scarce, posing an additional layer of complexity to an already challenging task. Limited work in this domain, especially for low-resource languages like Bengali, reveals a critical gap [21]. This paper addresses this critical gap by proposing a multi-domain dataset for GEE systems evaluation in Bengali. Unlike existing synthetic datasets [18], our dataset spans proficiency levels and diverse error types, incorporating real-world sentences from domains such as Bengali essays, social media, and news. Each of the 3402 sentences in the dataset features a single human-written correction, offering holistic fluency rewrites, thereby establishing a gold standard for the field to aspire to. The diversity in edits mirrors the challenges GEC systems encounter, emphasizing the need for context-aware explanations. The code and proposed dataset can be found at https://github.com/my625/bengali_gee. In summary, our contributions encompass two main aspects. *(i)* We present a real-world, multi-domain dataset of Bengali sentences with several grammatical error types, providing a robust evaluation framework. To the best of our knowledge, we are the *first* to curate such a real-world dataset where a single sentence may consist of multiple grammatical errors. *(ii)* Utilizing a two-step Bengali GEE pipeline, we assess the performance of various generative pre-trained LLMs compared to human experts. Furthermore, we evaluate the quality of the grammatical corrections using automated evaluation metrics such as precision, recall, $F_1$ score, $F_{0.5}$ score, and exact match (see Section 7.1). We also assess the quality of the explanations through human evaluation (see Section 7.2), shedding light on the capabilities of these LLMs in generating meaningful error explanations for the Bengali language.

## 2. DATASET

In the development of our dataset for GEE systems evaluation in Bengali, our primary objective was to overcome existing limitations, especially the absence of real-world, multi-error sentences. Drawing inspiration from the methodology of [25] and recognizing the inadequacies of minimal edits to capture fluency improvements, we curated a diverse corpus. Our dataset comprises 3402 Bengali sentences sourced from various domains, including essays written by school students with different proficiency levels, as well as content from social media and news (see Appendix A for dataset sources).

We collected 187 essays from 187 students enrolled in high school-level Bengali language courses in West Bengal, offering insight into grammatical errors made by students with diverse proficiency levels. The consent process for essay collection from students included providing clear explanations regarding the purpose of the data collection, the voluntary nature of participation, and the assurance of confidentiality and anonymity. The essays were collected from students who voluntarily agreed, with parental consent also obtained for those students. Subsequently, personal information in the collected essays was anonymized to ensure the privacy of the participants. Furthermore, to capture the dynamic nature of online communication, we crawled posts and comments from various public Bengali social media pages, including news outlets and community-driven platforms. Subsequently, the personal information in the crawled posts and comments was anonymized.

Each Bengali sentence in the dataset is paired with a single human-written correction, focusing on more comprehensive fluency revisions rather than minimal edits. This approach aimed to create a comprehensive dataset that authentically reflects the varied types of changes necessary for Bengali language learners and users. The entire annotation process has been carried out by three native Bengali language teachers (L1), who were appointed through Surge AI[1] and possess expertise in Bengali language. To ensure the quality and authenticity of the corrections, we engaged four other proficient Bengali language experts native to West Bengal and Bangladesh. They qualified through a correction task and contributed valuable annotations[2]. A blind test set was created for community evaluation by withdrawing half of the dataset from the analysis. The mean *levenshtein distance* (LD) between the original and corrected Bengali sentences was more than twice that of the existing corpora, emphasizing the substantial edits made to enhance fluency. It was observed that less grammatical Bengali sentences require more extensive changes and are inherently harder to correct. In addition, expert Bengali language L1 teachers also examined all sentence pairs in our dataset to categorize errors. The three cognitive levels [16], namely *single-word level errors*, *inter-word level errors*, and *discourse level errors*, contain various specific error categories. These categories guided the labeling of our dataset, including the corresponding types of correction applied. It is important to note that our curated dataset consists of sentences in which a single sentence may contain more than one type of error. This curated, multi-domain dataset not only serves as a gold standard for evaluating GEE systems but also addresses the crucial need for a real-world benchmark in Bengali GEC.

## 3. ERROR CATEGORIES WITH DATASET INSIGHTS

Following [16, 29, 28] and through consultation with human experts, we delineate the error categories within three cognitive levels as outlined below (see Appendix B for error category definition): *(i) Single-word level errors:* Errors at the single-word level occur in the initial and most basic cognitive stage, typically involving the improper use of *spelling* and *orthography*, often resulting from misremembering. 28.89% of the errors in our dataset belong to the *single-word level* error category. *(ii) Inter-word level errors:* These types of errors occur at the second cognitive level and typically arise from a misunderstanding of the target language, Bengali. This level is a common source of errors with evident manifestations as it involves the interaction between words. Bengali errors can be categorized into two linguistic classes: syntax and morphology. In terms of syntax, error types include *case markers*, *participles*, *subject-verb agreement*, *auxiliary verb*, *pronoun*, and *Guruchondali dosh*. For morphology, error types include *postpositions*. Our dataset contains 37.65% errors at the *inter-word level*. *(iii) Discourse level errors:* These types of errors represent the highest cognitive level, which requires a complete understanding of the context. Bengali discourse errors encompass issues such as *punctuation*, *verb tense*, *word order*, and *sentence structure*. In our dataset, 33.46% of errors are classified under the discourse level category. The dataset statistics, along with examples for a few dominant Bengali grammatical error types in the proposed corpus, are depicted in Table 1.

## 4. TASK FORMULATION

---

[1] Surge AI: https://www.surgehq.ai/faq

[2] We use Surge AI as our data annotation platform.

Table 1: Dataset statistics, including examples of a few dominant Bengali grammatical error types in the proposed corpus. Red text indicates erroneous words. Red underlined text denotes the erroneous word(s) containing the corresponding error type. Blue text indicates the corrected ones. Curly braces '{}' denotes a blank. Glosses of correct sentences are provided in round brackets '()'.

| Error Type | Percentage (%) | Examples |
|---|---|---|
| *Spelling* | 15.56 | [ধনি → ধনী]-দরিদ্র, পণ্ডিত-[মুর্খ → মূর্খ], শত্রু-মিত্র, সকলকে ভালোবাসা দরকার। (Gloss: Rich-poor, wise-foolish, enemy-friend, everyone love need.) |
| *Orthography* | 13.33 | শ্রদ্ধেয় শ্রী [মণি শঙ্কর → মণিশঙ্কর] [মুখোপাধ্যায় কে → মুখোপাধ্যায়কে] [জন্মদিনে → জন্মদিনের] শ্রদ্ধা ও প্রণাম [{} → জানাই] । (Gloss: Respected to Mr. Manishankar Mukhopadhyay of birthday respect and reverence convey.) |
| *Case-marker* | 7.51 | [ছাত্রটি → ছাত্রটির] [কির্তী → কীর্তি] [সবাই → সবাইকে] অবাক করেছিল। (Gloss: Of the student achievement everyone amazed did.) |
| *Subject-verb agreement* | 10.23 | [তাঁরা → তারা] এখন [অনুষ্ঠান → অনুষ্ঠানে] যোগ দিতে [যাচ্ছি → যাচ্ছে] । (Gloss: They now to the event participation to give are going.) |
| *Auxiliary verbs* | 5.89 | দ্রুতগতি সম্পন্ন ঘোড়ার পিঠে ছুটে [{} → গিয়ে] তারা চারদিক থেকে ঘিরে ফেললো [প্রতিপক্ষকে → প্রতিপক্ষকে] । (Gloss: Fast-paced equipped with of the horse on the back by running they from all directions surrounded the opponents.) |
| *Pronoun* | 4.23 | [তাঁরা → তারা] এখন [অনুষ্ঠান → অনুষ্ঠানে] যোগ দিতে [যাচ্ছি → যাচ্ছে] । (Gloss: They now to the event participation to give are going.) |
| *Guruchondali dosh* | 8.10 | [সন্ধ্যা → সন্ধ্যার] অন্ধকার নামিয়া [এসেছে → আসিয়াছে] । (Gloss: Of the evening darkness having descended has come.) |
| *Punctuation* | 10.91 | আমি এই ক্লাস করতে চাই [{} → ,] স্যার [{} → ।] কিভাবে হবে [। → ?] (Gloss: I this class want to do, Sir. How will be?) |
| *Verb tense* | 9.15 | আমি গতকাল বাজারে [যাই → গিয়ে] সবজি [কিনি → কিনলাম] । (Gloss: I yesterday to the market went vegetables bought.) |
| *Word order* | 10.22 | ছাত্রদের [উচিত → বুঝতে] শেখা [বুঝতে → উচিত] তাদের একটা কর্তব্য আছে । (Gloss: Students to understand to learn necessary their one duty is.) |

Given the input to the GEE model (LLM) as an erroneous Bengali sentence $S_{\text{err}} = \{w_1, w_2, w_3, \ldots, w_n\}$, the task involves two main components:

*(1) Produce the corrected sentence ($S_{corr}$):* Identify and rectify errors in the provided sentence $S_{\text{err}}$ to create a corrected sentence $S_{\text{corr}} = \{w'_1, w'_2, w'_3, \ldots, w'_m\}$ that is grammatically and contextually accurate in Bengali.

$$S_{\text{corr}} = \text{LLM}(S_{\text{err}})$$

*(2) Provide concise explanations for each error type:* Categorize each corrected error in $S_{\text{corr}}$ into specific error types and offer a brief explanation for each type of error. Clarify the *grammatical*, *syntactical*, or *semantic issues* addressed, and present the rationale behind each correction.

$$E_{\text{types}} = \text{LLM}(S_{\text{err}}, S_{\text{corr}})$$

where $E_{\text{types}}$ is a set of error types corresponding to each corrected error in $S_{\text{corr}}$.

The goal of this task is to enhance understanding of the language intricacies involved in error correction for Bengali sentences. The corrected sentence $S_{\text{corr}}$ and their associated error type explanations $E_{\text{types}}$ serve as valuable resources to improve automatic error correction systems in the Bengali language.

## 5. METHODOLOGY

In conducting our GEE task, we prompt generative pre-trained LLMs in one-shot mode, employing a comprehensive methodology that leverages the capabilities of various LLMs. Specifically, we prompted GPT-4 Turbo (GPT-4), GPT-3.5 Turbo (GPT-3.5), Text-davinci-003 (Davinci), Text-babbage-001 (Babbage), Text-curie-001 (Curie), and Text-ada-001 (Ada) through the OpenAI API[3], as well as Llama-2-7b, Llama-2-13b, and Llama-2-70b via the Llama-2-Chat API[4]. Our systematic experimental process involved both LLM and human experts, independently performing two crucial tasks. *First*, they were tasked with *producing the corrected sentence* in Bengali by identifying and rectifying errors in the provided sentences, ensuring grammatical and contextual accuracy. *Second*, for every corrected error, they were required to categorize the *type of error* and provide concise *explanations* on the grammatical, syntactical, or semantic problems addressed. The one-shot prompt used for the Bengali grammatical error explanation task is shown in Figure 2. We also investigated various alternative few-shot prompting techniques, elaborated further in Appendix C. By employing this multifaceted methodology, our objective was to holistically assess the relative performance of each LLM, evaluating their proficiency in generating the *corrected sentence* and providing *concise explanations* for each error type. Furthermore, we compared the capabilities of LLMs with human experts[5] (baseline). Each LLM assessed all sentences, while the human experts collectively assessed the entire set of sentences. For the human experts, the entire set of erroneous sentences was divided into four parts, with each expert assessing their assigned portion. It should be noted that to ensure a fair and meaningful comparison, we present the same prompt format to both LLMs and human experts, as shown in Figure 2.

## 6. EVALUATION CRITERIA

Following [16], the *automatic evaluation* process encompasses scrutiny at both the token level (precision, recall, $F_1$, and $F_{0.5}$) and the sentence level (exact match). The *exact match* (EM) evaluates the agreement between the predicted and gold standard sentences. Assessing the quality of *explanations* presents challenges due to the potential for multiple ways of explaining errors. Achieving reliable automatic evaluation in GEE for Bengali requires multi-reference metrics such as METEOR [2] and benchmarks with multiple references for each error; however, creating such datasets is resource-intensive. We engaged three other experienced Bengali language teachers, who have expertise in Bengali language, through UpWork[6] to assess the explanations (i.e., *human evaluation*), as detailed in Section 7.2. Given their expertise in teaching Bengali, they can provide reliable judgments on the correctness and informativeness of explanations while considering the identified error types.

## 7. EXPERIMENTAL RESULTS

This section presents the results of the automatic and human evaluation of Bengali GEE.

---

[3] OpenAI: https://platform.openai.com/docs/models/
[4] Llama 2: https://huggingface.co/TheBloke
[5] We hired four Bengali language teachers through the UpWork online platform, each possessing extensive experience in teaching the Bengali language.
[6] UpWork: https://www.upwork.com/

```
You have been given Bengali sentence(s) with errors.  Your assignment has two main components:

(1) Produce the Corrected Sentence:
Identify and rectify the errors in the provided sentence, ensuring it is both grammatically and contextually accurate in Bengali.

(2) Provide Concise Explanations for Each Error Type:
For every error corrected in the sentence, categorize the error type and offer a brief explanation.  Clarify the grammatical, syntactical, or semantic issues
addressed and present the rationale behind each correction.  The goal is to enhance understanding of the language intricacies involved.

Example:

Incorrect Sentence:
আমি গতকাল বাজারে যাই সবজি কিনি। (Gloss:  I yesterday to the market go vegetables buy.)

Corrected Sentence:
আমি গতকাল বাজারে গিয়ে সবজি কিনলাম। (Gloss:  I yesterday to the market went vegetables bought.)

Explanations:

1. যাই (Gloss:  Go)  → গিয়ে (Gloss:  Went)
Error Type:  Verb Tense
Explanation:  The verb "যাই" (Gloss:  Go) implies the act of going, while the context of the sentence requires the past action of going to the market.  There-
fore, the correct form is "গিয়ে" (Gloss:  Went), indicating that the speaker went to the market.

2. কিনি (Gloss:  Buy)  → কিনলাম (Gloss:  Bought)
Error Type:  Verb Tense
Explanation:  The original sentence used the verb "কিনি" (Gloss:  Buy), which is the present tense form.  To match the past tense context of the sentence, the
correct form is "কিনলাম" (Gloss:  Bought), indicating that the speaker bought vegetables in the past.

Sentence(s) for correction is/are provided below.

{Insert the Bengali sentence(s) here}
```

Figure 2: One-shot prompt used for the Bengali GEE task. Glosses are provided in round brackets '()'.

## 7.1 Automatic Evaluation

In this section, we present a performance comparison for predicting grammatically correct Bengali sentences, considering various types of errors in the proposed corpus. The comparison is conducted between human experts (baseline) and different generative pre-trained LLMs. As shown in Table 2, GPT-4 Turbo consistently demonstrates better agreement with human experts across different error types compared to other LLMs, demonstrating its superior performance. On the other hand, Text-ada-001 exhibits the most substantial deviation from human experts, with consistently lower metrics in precision, recall, $F_1$ score, $F_{0.5}$ score, and exact match (EM) across all error levels (i.e., *single-word level*, *inter-word level*, and *discourse level*). Following [9], to determine the alignment between the best-performing LLM (i.e., GPT-4 Turbo) and human experts in terms of various automated evaluation metrics, we calculate the Pearson correlation coefficient ($r$) [17] between the two. Table 3 indicates that $F_{0.5}$ achieves the highest correlation score. Additionally, precision demonstrates a stronger correlation with human experts compared to recall. However, human experts consistently outperform LLMs and achieve the best results in predicting grammatically correct Bengali sentences across all automated metrics for various types of errors.

## 7.2 Human Evaluation

As shown in Table 2, GPT-4 Turbo exhibited better results in predicting grammatically correct Bengali sentences compared to other LLMs. We compared various Bengali grammatical error explanations provided by GPT-4 Turbo and human experts. For each erroneous sentence, we present the corrected sentence and explanations generated by GPT-4 Turbo and a human expert to one of three teachers (as discussed in Section 6). They are asked to check for two types of mistakes[7] in the explanations: ***wrong error type*** (an error type that is not present in the erroneous sentence according to the gold standard error type) and ***wrong error explanation*** (an error explanation that is not present for the particular error type provided by human experts). Table 4 shows that GPT-4 Turbo generates 27.32% *wrong error type* and 35.89% *wrong error explanation*.

---

[7]We refer to grammatical errors in sentences as *errors*, and errors made by LLMs are termed as *mistakes*.

As depicted in Table 5, during the assessment of the erroneous Bengali sentence "সন্ধ্যা অন্ধকার নামিয়া এসেছে।" (Gloss: Evening darkness having descended has come.), GPT-4 Turbo offered corrections and explanations. However, compared to the corrections made by a human expert, several notable shortcomings in GPT-4 Turbo's predictions became evident. The primary discrepancy lies in GPT-4 Turbo's choice to correct "নামিয়া" (Gloss: Having descended) to "নামিয়ে" (Gloss: Descending). GPT-4 Turbo categorized this as a *verb form* error, indicating that the original sentence had a *spelling* mistake. However, the human expert correctly identified that "নামিয়া" (Gloss: Having descended) was the accurate term. GPT-4 Turbo's suggested term "নামিয়ে" (Gloss: Descending) does not accurately capture the intended action and meaning of the original sentence. This highlights a limitation in GPT-4 Turbo's understanding of contextual nuances and specific *verb* forms in Bengali. Furthermore, GPT-4 Turbo failed to address the human expert's correction regarding the *case-marker* for "সন্ধ্যা" (Gloss: Evening). The human expert rightly pointed out that "সন্ধ্যা" (Gloss: Evening) should be corrected to "সন্ধ্যার" (Gloss: *Of* the evening), adding the *case-marker* "র" (Gloss: *Of*) to indicate possession or association with the evening darkness. This detail was overlooked by GPT-4 Turbo, indicating a lack of attention to grammatical nuances and case-marking rules in Bengali. Furthermore, GPT-4 Turbo did not consider the *Guruchondali dosh* in the *verb* form. It was not able to identify the need for a change from "এসেছে" (Gloss: Has come) to "আসিয়াছে" (Gloss: Has come), and thus the explanation did not address the agreement between the *verb* form and the action of "নামিয়া" (Gloss: Having descended). On the other hand, the human expert accurately identified and addressed this linguistic nuance, ensuring a more contextually appropriate correction. In summary, the human evaluation of GPT-4 Turbo in correcting Bengali sentences revealed consistent limitations. GPT-4 Turbo struggled with nuanced aspects such as *word order* errors, *Guruchondali dosh*, *case marker* errors, *spelling* errors, etc. In particular, GPT-4 Turbo fell short in capturing the intricacies of Bengali language subtleties, hindering its ability to provide satisfactory explanations for error types. Furthermore, the model demonstrated ineffectiveness even with short-length Bengali sentences containing multiple grammatical errors (see Table 5). In contrast, the human expert's corrections and detailed explanations consistently showcased a deeper understanding of these nuances. This underscores the imperative for human oversight in explaining Bengali grammatical errors, particularly when addressing complex linguistic intricacies and context-specific meanings.

Table 2: Performance comparison in predicting grammatically correct Bengali sentences for various error types and overall on the proposed corpus between human experts (baseline) and LLMs. EM denotes *exact match*.

| Metric | Human | GPT-4 | GPT-3.5 | Llama-2-70b | Llama-2-13b | Llama-2-7b | Davinci | Curie | Babbage | Ada |
|---|---|---|---|---|---|---|---|---|---|---|
| *Single-word level errors* | | | | | | | | | | |
| Precision | 95.89 | 74.47 | 69.90 | 71.84 | 68.81 | 65.62 | 67.82 | 63.69 | 62.91 | 60.41 |
| Recall | 90.82 | 72.81 | 66.81 | 68.90 | 64.91 | 62.59 | 63.91 | 62.36 | 60.75 | 57.28 |
| $F_1$ | 92.47 | 73.39 | 67.35 | 69.32 | 66.72 | 63.49 | 65.11 | 62.89 | 61.29 | 58.95 |
| $F_{0.5}$ | 94.91 | 73.81 | 68.79 | 70.61 | 67.17 | 64.91 | 66.90 | 63.91 | 62.89 | 59.96 |
| EM | 73.56 | 48.69 | 39.62 | 45.30 | 43.74 | 41.29 | 46.72 | 42.89 | 40.91 | 37.53 |
| *Inter-word level errors* | | | | | | | | | | |
| Precision | 90.48 | 68.84 | 62.91 | 63.72 | 60.47 | 58.67 | 60.41 | 58.73 | 54.71 | 53.04 |
| Recall | 87.93 | 65.60 | 60.74 | 60.73 | 56.83 | 54.67 | 57.48 | 54.38 | 50.75 | 50.49 |
| $F_1$ | 88.20 | 66.39 | 61.35 | 61.49 | 58.72 | 56.43 | 58.18 | 56.82 | 52.73 | 51.06 |
| $F_{0.5}$ | 89.73 | 67.82 | 62.35 | 62.28 | 59.49 | 57.48 | 59.39 | 57.92 | 53.42 | 52.48 |
| EM | 69.46 | 46.70 | 43.91 | 45.80 | 42.85 | 40.91 | 43.79 | 41.84 | 39.90 | 38.61 |
| *Discourse level errors* | | | | | | | | | | |
| Precision | 94.52 | 70.57 | 67.88 | 65.84 | 63.74 | 62.72 | 63.71 | 60.15 | 58.27 | 56.41 |
| Recall | 89.82 | 67.75 | 65.81 | 62.83 | 61.56 | 60.48 | 60.92 | 56.41 | 55.75 | 54.28 |
| $F_1$ | 91.62 | 69.42 | 66.32 | 63.75 | 62.27 | 61.71 | 61.11 | 58.93 | 56.29 | 55.95 |
| $F_{0.5}$ | 92.47 | 70.47 | 66.74 | 64.11 | 63.19 | 62.31 | 62.90 | 59.74 | 57.73 | 55.99 |
| EM | 72.79 | 50.71 | 46.83 | 45.92 | 43.88 | 40.54 | 44.98 | 42.89 | 38.91 | 37.61 |
| *Overall* | | | | | | | | | | |
| Precision | 93.55 | 71.11 | 66.79 | 66.85 | 64.10 | 62.19 | 63.78 | 60.63 | 58.41 | 56.43 |
| Recall | 89.46 | 68.48 | 64.39 | 63.87 | 60.99 | 57.35 | 59.14 | 57.35 | 55.51 | 55.51 |
| $F_1$ | 90.74 | 69.54 | 64.94 | 64.59 | 62.36 | 61.03 | 60.45 | 59.36 | 56.53 | 55.87 |
| $F_{0.5}$ | 92.24 | 70.54 | 65.85 | 65.36 | 63.09 | 61.43 | 61.73 | 60.87 | 57.77 | 55.95 |
| EM | 70.60 | 49.04 | 43.89 | 45.69 | 43.44 | 40.89 | 45.03 | 42.49 | 39.95 | 37.95 |

Table 3: The Pearson correlation coefficient ($r$) between the best-performing LLM (i.e., GPT-4 Turbo) and human experts in terms of various automated evaluation metrics. EM denotes *exact match*.

| Precision | Recall | $F_1$ | $F_{0.5}$ | EM |
|---|---|---|---|---|
| 0.485 | 0.434 | 0.471 | **0.497** | 0.456 |

Table 4: Results of human evaluation of different LLMs in Bengali GEE. WET denotes *wrong error type* and WEE denotes *wrong error explanation*.

| Metric | GPT-4 | GPT-3.5 | Llama-2-70b | Llama-2-13b | Llama-2-7b | Davinci | Curie | Babbage | Ada |
|---|---|---|---|---|---|---|---|---|---|
| WET (%) | 27.32 | 30.37 | 33.19 | 35.20 | 37.84 | 32.05 | 35.35 | 39.51 | 42.92 |
| WEE (%) | 35.89 | 38.82 | 39.04 | 40.48 | 45.82 | 38.98 | 42.27 | 45.95 | 49.87 |

## 8. RELATED WORK

Despite the growing interest in GEC and the availability of GEC datasets in high-resource languages such as English [11, 6, 20], Chinese [32], German [4], Russian [24], Spanish [12], etc., there is a noticeable scarcity of real-world GEC datasets specifically tailored for low-resource languages such as Bengali. Although there is existing GEC research for Bengali [23, 27, 19, 18, 1], no work has been undertaken in the areas of feedback or explanation generation within this context. A notable effort by [10] in feedback comment generation (FCG) introduces a typology for learning feedback, covering abstract types (e.g., *tone* and *idiom*) and grammatical pattern types (e.g., *comparative* and *causative*) in the English language. However, their work is in an early stage, lacking human or automatic evaluation of comment quality. Our GEE task addresses this gap by emphasizing the explicit communication of grammatical rules and linguistic insights in error correction, followed by automatic and human evaluation for Bengali.

## 9. CONCLUSION

We explore GEE specifically for the Bengali language. The objective of GEE is to enhance user understanding and engagement with error correction systems, providing comprehensive insight into language nuances and opportunities for improvement. We present a real-world multi-domain dataset for Bengali proficiency evaluation in GEE systems, comprising 3402 sentences from various domains such as Bengali essays, social media, and news. Through assessing various generative pre-trained LLMs and comparing their performance with human experts, we observe that GPT-4 Turbo performs comparatively better than other LLMs but faces challenges in nuanced aspects such as *word-order error*, *spelling error*, *case-marker error*, *Guruchondali dosh*, etc. GPT-4 Turbo's limitations are evident even with short-length sentences containing multiple errors. In contrast, human experts consistently surpass the LLMs, highlighting the necessity of human oversight in the task of explaining Bengali grammatical errors. In scaling this approach beyond the confines of this paper, collaboration with Bengali language instructors and educators presents a promising avenue.

Table 5: An example (containing *case-marker error* and *Guruchondali dosh*) showcasing `GPT-4 Turbo`'s GEE output utilizing a one-shot prompt and presenting a comparison with a human expert. Glosses are provided in round brackets '()'.

---

**User Input**: সন্ধ্যা অন্ধকার নামিয়া এসেছে। (Gloss: Evening darkness having descended has come.)

**GPT-4 Turbo**

**Corrected Sentence**: সন্ধ্যা অন্ধকার নামিয়ে এসেছে। (Gloss: Evening darkness descending has come.)

**Explanations**:
1. নামিয়া (Gloss: Having descended) → নামিয়ে (Gloss: Descending)

**Error Type**: Verb Form

**Explanation**: The original sentence had a spelling mistake. The correct form is "নামিয়ে" (Gloss: Descending), indicating the act of coming. The verb "এসেছে" (Gloss: Has come) requires the addition of the appropriate verb form, and "নামিয়ে" (Gloss: Descending) complements it correctly in this context.

**Human**

**Corrected Sentence**: সন্ধ্যার অন্ধকার নামিয়া আসিয়াছে। (Gloss: Of the evening darkness having descended has come.)

**Explanations**:
1. সন্ধ্যা (Gloss: Evening) → সন্ধ্যার (Gloss: Of the evening)

**Error Type**: Case-marker

**Explanation**: In the original sentence, "সন্ধ্যা" (Gloss: Evening), the word lacks the appropriate case-marker. Bengali often uses case-markers to indicate possession or association. In this context, the correct form is "সন্ধ্যার" (Gloss: Of the evening), where "র" (Gloss: Of) is added as a case-marker to signify that it is the darkness of the evening. Therefore, the corrected version is "সন্ধ্যার অন্ধকার" (Gloss: Of the evening darkness).

2. এসেছে (Gloss: Has come) → আসিয়াছে (Gloss: Has come)

**Error Type**: Guruchondali dosh

**Explanation**: The original sentence uses "এসেছে" (Gloss: Has come) to convey the arrival of darkness. However, to agree with the action indicated by the word "নামিয়া" (Gloss: Having descended), the correct form for this context is "আসিয়াছে" (Gloss: Has come). The verb "আসিয়াছে" (Gloss: Has come) is more appropriate to describe the evening darkness descending or arriving. This correction specifically addresses the Guruchondali dosh, ensuring the use of the appropriate verb form in agreement with the action of "নামিয়া" (Gloss: Having descended).

# APPENDIX
## A. DATASET SOURCES

We curated the essay dataset from expert-verified erroneous Bengali sentences. These were sourced from the answer scripts of the final exams of the tenth-standard students from a high school in West Bengal, totaling 1678 sentences. Furthermore, we collected 1724 expert-verified erroneous Bengali sentences from crawled posts and comments on various public Bengali social media pages, including news outlets and community-driven pages. Specifically, the dataset includes content sourced from the Facebook pages such as:

- `https://www.facebook.com/BanglaNEWJ/`
- `https://www.facebook.com/eisamay.com/`
- `https://www.facebook.com/livecalcuttanews`
- `https://www.facebook.com/News18Bangla`
- `https://www.facebook.com/abpananda`
- `https://www.facebook.com/RepublicBangla/`
- `https://www.facebook.com/BBCBengaliService/`
- `https://www.facebook.com/AnandabazarSocial`

This multi-source approach ensures the inclusion of diverse linguistic contexts. To guarantee the reliability of the dataset, we employed proficient, experienced, and educated native Bengali language teachers to filter out erroneous sentences, thus improving the quality and authenticity of the dataset for the evaluation of the GEE system.

## B. CATEGORIES OF GRAMMATICAL ERRORS

The description for each grammatical error category in our proposed dataset is presented as follows:

- *Spelling*: A spelling error in Bengali grammar is characterized by inaccuracies or deviations in the correct formation of words or characters within the Bengali language. These errors specifically pertain to incorrect arrangement of letters or characters, resulting in discrepancies from established spelling norms and accepted conventions in Bengali writing.

- *Orthography*: An orthographic error in Bengali grammar refers to inaccuracies or deviations from the accepted conventions of spelling and writing in the Bengali language. These errors include misalignments in the arrangement and representation of letters, words, or characters, leading to a departure from the standard orthographic rules of Bengali. Unlike spelling errors, orthographic errors extend beyond individual word formations, addressing broader issues in the overall structure and presentation of written Bengali text.

- *Case marker*: A case marker error in Bengali grammar occurs when there are inaccuracies in applying the appropriate case markers to nouns and pronouns. This can happen in various instances, such as using non-standard markers like -রে (-re) instead of the standard -কে (-ke) in certain dialects for the objective case. Genitive case errors may occur if the endings -এর (-er) or -র (-r) are incorrectly applied based on the characteristics of the noun. Similarly, locative case errors arise when incorrect markers (ে/-e, -তে/-te, or -য়/-y) are used, deviating from the grammatical rules.

- *Participles*: In Bengali grammar, a participle error occurs when there are inaccuracies in the usage of particiles, which can sometimes be confused with ordinary verbs like present simple, past simple, or present continuous. Errors can also arise in situations related to other participants, affecting the overall structure and clarity of the sentence.

- *Subject-verb agreement*: This type of error occurs when there is a discrepancy in person and number between the subject and the verb. In Bengali grammar, person denotes the grammatical category indicating the speaker's relationship with the subject. Bengali grammar recognizes three persons: first person, second person, and third person. To ensure proper grammar in Bengali, the verb must align with the subject in both person and number. For example, if the subject is singular, the verb should also be singular, and if the subject is plural, the verb should be plural. Likewise, if the subject is in the first person, the verb should also be in the first person, and so forth.

- *Auxiliary verb*: An auxiliary verb error in Bengali grammar occurs when a sentence lacks the necessary auxiliary verb. Auxiliary verbs are essential to convey a tense, mood, or voice. Errors in the usage of auxiliary verbs can lead to incomplete or unclear expressions, which can impact the overall grammatical structure and communicative effectiveness of the sentence.
- *Pronoun*: Pronoun errors in Bengali can occur due to inaccuracies in personal pronoun usage. Although Bengali pronouns are similar to English, distinctions in addressing first, second, and third persons, as well as singular and plural forms, may lead to challenges. Issues can arise in using third-person pronouns that accurately indicate proximity. Errors may also arise when using second and third-person pronouns with varying forms for familiarity and politeness.
- *Guruchondali dosh*: Guruchondali dosh in Bengali grammar signifies a linguistic error arising from the inappropriate blending of *sadhu bhasha*, the formal literary style, and *cholito bhasha*, the informal colloquial style. This grammatical infraction is considered when these two distinct styles are mixed within a written document. *Sadhu bhasha* maintains a formal and refined structure, while *cholito bhasha* represents the informal spoken language. The error results in a lack of linguistic coherence, making the text difficult for the reader to understand because of the conflicting nature of the formal and informal elements.
- *Postpositions*: In Bengali grammar, there is a difference in the use of prepositions compared to English. Although English employs prepositions that appear before their objects (e.g., "*beside* him", "*inside* the house"), Bengali typically uses postpositions that come after their objects (e.g., "*or pashe*", "*bair bhitore*"). A postposition error in Bengali grammar occurs when there are inaccuracies in the usage of postpositions, which typically come after their objects. Some postpositions in Bengali require that their object nouns take the genitive case, while others require the objective case, and it is important to memorize this distinction. The majority of postpositions are created by taking nouns related to a location and inflecting them for the locative case, and they can also be applied to verbal nouns.
- *Punctuation*: Punctuation symbols serve to elucidate the meaning and structure of a sentence in writing. In Bengali, familiar punctuation marks encompass the period (।), comma (,), semicolon (;), colon (:), question mark (?), and exclamation mark (!). An error of this kind arises when these punctuation marks are either omitted or used incorrectly within a sentence.
- *Verb tense*: A verb tense error in Bengali grammar occurs when there is a discrepancy in the accurate expression of the temporal relationship between the subject and the verb. Bengali grammar encompasses various tenses, such as past, present, and future, which convey the timing of actions or events. A verb tense error arises when the chosen tense of the verb does not correspond correctly with the intended time frame or when there is inconsistency in the use of tenses within a sentence or paragraph.
- *Word order*: A word order error in Bengali grammar occurs when there are deviations from the standard arrangement of words in a sentence. Bengali typically follows a subject-object-verb (SOV) word order, where the subject precedes the object, and the verb appears at the end of the sentence. Errors arise when words are misplaced, disrupting the established structure.
- *Sentence structure*: This occurs when the structure of a sentence is flawed, leading to grammatical inaccuracies that can alter the intended meaning and make the sentence difficult to comprehend.

## C. FEW-SHOT PROMPTING

We also examined various other few-shot prompting methods, including `two-shot`, `four-shot`, `eight-shot`, and `sixteen-shot` examples. Our focus was on investigating the few-shot prompting approach using GPT-4 Turbo, given its superior performance among LLMs. Despite our extensive exploration, the results of the few-shot prompting experiments did not show significant improvements across the evaluated criteria. Consequently, the outcomes obtained from the few-shot prompting approach are not included in this paper.